\documentclass[journal]{IEEEtran}
\usepackage{xcolor}
\usepackage{graphicx}

\usepackage{amsmath}
\usepackage{amssymb}
\usepackage{url}
\usepackage[pagebackref=false,breaklinks=true,letterpaper=true,colorlinks,bookmarks=false]{hyperref}
\usepackage{caption}
\usepackage{xcolor}
\newcommand{\todo}[1]{}
\renewcommand{\todo}[1]{{\color{red} \textbf{{#1}}}}

\begin{document}
\title{Simultaneous imputation and disease classification in incomplete medical datasets using Multigraph Geometric Matrix Completion (MGMC)}

\author{Gerome~Vivar,
        Anees~Kazi,
        Hendrik~Burwinkel,
        Andreas~Zwergal,
        Nassir~Navab,
        Seyed-Ahmad~Ahmadi$^\dagger$, \\for the Parkinson’s Progression Markers and Alzheimer's Disease Neuroimaging Initiatives$^*$
\thanks{G. Vivar, A. Kazi, H. Burwinkel, N. Navab and S.-A. Ahmadi are with the Department
of Computer Aided Medical Procedures (CAMP), Technical University of Munich (TUM), Boltzmannstr. 3, 85748 Garching, Germany.}
\thanks{G. Vivar, A. Zwergal and S-A. Ahmadi are with the German Center for Vertigo and Balance Disorders (DSGZ), Ludwig-Maximilians University (LMU), Marchioninistr. 15, 81377 Munich, Germany.}
\thanks{$^\dagger$S-A.A. is the corresponding author (e-mail: \href{mailto:ahmadi@cs.tum.edu}{ahmadi@cs.tum.edu})}
\thanks{This work was supported by the German Federal Ministry of Education and Health (BMBF) in connection with the foundation of the German Center for Vertigo and Balance Disorders (DSGZ) [grant 01 EO 0901], with partial support of "Freunde und F{\"o}rderer der Augenklinik M{\"u}nchen", Germany.}
\thanks{$^*$Data used in preparation of this article were obtained from the Alzheimer's Disease Neuroimaging Initiative (ADNI) database (\url{adni.loni.usc.edu}). As such, the investigators within the ADNI contributed to the design and implementation of ADNI and/or provided data but did not participate in analysis or writing of this report. A complete listing of ADNI investigators can be found at: \url{http://adni.loni.usc.edu/wp-content/uploads/how_to_apply/ADNI_Acknowledgement_List.pdf}. Further data were obtained from the Parkinson’s Progression
Markers Initiative (PPMI) database \url{www.ppmi-info.org/data}. List of all PPMI funding partners can be found at \url{www.ppmi-info.org/fundingpartners}.}}

\markboth{Vivar et al., Simultaneous imputation and disease classification using MGMC.}%
{Shell \MakeLowercase{\textit{et al.}}: Bare Demo of IEEEtran.cls for IEEE Journals}

\maketitle
\begin{abstract} 
Large-scale population-based studies in medicine are a key resource towards better diagnosis, monitoring, and treatment of diseases. They also serve as enablers of clinical decision support systems, in particular Computer Aided Diagnosis (CADx) using machine learning (ML). Numerous ML approaches for CADx have been proposed in literature. However, these approaches assume full data availability, which is not always feasible in clinical data. To account for missing data, incomplete data samples are either removed or imputed, which could lead to data bias and may negatively affect classification performance. As a solution, we propose an end-to-end learning of imputation and disease prediction of incomplete medical datasets via Multi-graph Geometric Matrix Completion (MGMC). MGMC uses multiple recurrent graph convolutional networks, where each graph represents an independent population model based on a key clinical meta-feature like age, sex, or cognitive function. Graph signal aggregation from local patient neighborhoods, combined with multi-graph signal fusion via self-attention, has a regularizing effect on both matrix reconstruction and classification performance. Our proposed approach is able to impute class relevant features as well as perform accurate classification on two publicly available medical datasets. We empirically show the superiority of our proposed approach in terms of classification and imputation performance when compared with state-of-the-art approaches. MGMC enables disease prediction in multi-modal and incomplete medical datasets. These findings could serve as baseline for future CADx approaches which utilize incomplete datasets.
\end{abstract}


%
\IEEEpeerreviewmaketitle

\section{Introduction}
\label{sec:intro}
\IEEEPARstart{L}{arge} population-based studies in medicine, acquired at multiple institutions, are instrumental resources for a better clinical understanding of the diagnosis, progression and treatment of diseases. In medical health informatics, they serve as fundamental enablers for the design and analysis of novel clinical decision support systems (CDSS) and Computer Aided Diagnosis (CADx) \cite{BHI:Gamache18, BHI:Machluf17, BHI:Dash19}. Often, such datasets incorporate multimodal data (imaging and non-imaging), in order to capture as many aspects of the disease as possible. Another characteristic is the occurrence of missing data, which is difficult to prevent \cite{IMPUTATION:Little12}. Missingness occurs both on a feature-level, e.g. after outlier removal due to low-quality data acquisition, as well as on a modality-level, e.g. when an examination was too costly or not sufficiently indicated. 

Two prominent examples for such datasets in neurology and neuroscience were published by the Alzheimer's Disease (AD) Neuroimaging Initiative (ADNI) \cite{DATASET:marinescu2018tadpole} and the Parkinson's disease (PD) Progressive Marker Initiative (PPMI) \cite{DATASET:marek2011parkinson}. Together, AD and PD are the most common neurodegenerative diseases, with AD accounting for 60-80\% of dementia cases, and PD affecting 1–2\% of the global population over the age of 65. Neurodegenerative diseases result in a progressive decay and death of nerve cells \cite{AD:Facts}. 
Increasing rates of up to a million new AD cases per year \cite{AlzheimersAssociation2014}, and the prospect of novel neuroprotective and disease-modifying therapeutics, in both AD \cite{NEURO:Gozes17neuroprotectionAD} and PD \cite{NEURO:Kim17neuroprotectionPD}, motivate an early diagnosis of these diseases, ideally already at a pre-symptomatic stage.

\textit{Related works:} 
Among ML-based CADx approaches, Thung et al. \cite{SWMC:Thung16, MedIA2018:Thung2018} use Low-Rank Matrix Completion (LRMC) to predict conversion of the disease in patients with Mild Cognitive Impairment (MCI) to Alzheimer’s Disease (AD). 
Zhou et al. \cite{TMI2019:Zhou2019} proposed to solve AD diagnosis using latent representation learning, by projecting both complete and incomplete modalities onto a common subspace. 
Both approaches by \cite{MedIA2018:Thung2018} and \cite{TMI2019:Zhou2019} assume a linear relationship between the input features and the target variable, and latent embeddings and linear classification are trained in two separate steps  \cite{TMI2019:Zhou2019}, which does not take advantage of end-to-end learning. 

A recent development in non-linear signal processing on unstructured domains is geometric deep learning \cite{GCN:Bronstein17}. A breakthrough paper in this domain was the first demonstration of semi-supervised classification using graph convolutional neural networks (GCN) \cite{GCN:Kipf16}. GCNs solve classification by learning optimal filters in the graph spectral domain. These filters perform non-linear message passing between vertices, in order to map high-dimensional signals to labels for all vertices in the graph. Parisot et al. \cite{GCN:Parisot18MedIA} transferred this approach to CADx in medicine, by introducing a novel concept for modeling patient populations as a graph: patient meta-data like demographics (e.g. sex, age, etc.) are used to compute similarities between patients, leading to an adjacency matrix with an associated graph Laplacian. Intuitively, the graph then represents the "social network" of patients in the cohort. As has been shown in several works, \cite{GCN:Parisot17, GCN:Parisot18MedIA, ISBI:Kazi19SelfAtt, MICCAI2019:Kazi, IPMI:Kazi19InceptionGCN, Coates2019}, GCNs can significantly improve the accuracy of CADx in medicine. Importantly, the CADx accuracy crucially depends on a meaningful graph representation of the cohort. To make GCNs more robust towards the patient similarity measures and graph construction, we have shown previously that it is beneficial to construct multiple graphs, one for each meta-feature, and fuse their processed features towards the decision layer, e.g. through attention mechanisms  \cite{ISBI:Kazi19SelfAtt, MICCAI2019:Kazi}. 

Regarding incomplete datasets, Monti et al. \cite{RGCN:Monti17} showed that geometric deep learning provides a principled framework for non-linear imputation, through geometric matrix completion (GMC) using GCNs and LSTMs. The target application were recommender systems, where a signal matrix denotes how users (rows) rank different items (columns). Naturally, user ratings are highly incomplete, and as such the signal matrix is sparsely populated. Through meta-data, two graphs can be constructed which represent the social network between users (row graph) and the semantic similarities between items (column graph). In GMC, the incomplete matrix and the two graphs serve as input to a GCN, which learns appropriate filters for non-linear signal diffusion on the graph domains, such that the entire completed matrix is reconstructed at the output. Essentially, the method learns to reconstruct the known entries in the sparse feature matrix, and fills missing entries as a side effect, purely through non-linear  global signal diffusion.

In our own previous work \cite{Vivar2018}, we utilized multi-target training to combine GMC with supervised classification into a Recurrent Graph Convolutional Network (RGCN). Similar to Parisot et al. \cite{GCN:Parisot18MedIA}, we constructed a patient graph from clinically relevant meta-data (e.g. age and sex of patients). We concatenated the incomplete feature matrix and incomplete labels, and trained a GCN for signal diffusion, along with a Long-Short Term Memory (LSTM) network for iterative matrix reconstruction. Both GCN and LSTM were trained end-to-end towards MCI to AD conversion prediction, with two weighted losses for simultaneous classification and imputation. 



\textit{Proposed approach:} Building up on our previous works in \cite{Vivar2018, MICCAI2019:Kazi}, we propose to solve disease classification in multimodal and incomplete datasets using Multi-graph Geometric Matrix Completion (MGMC). The contributions of this work are threefold: 1) we formulate the disease classification problem in multimodal and incomplete datasets using MGMC; 2), we propose a novel method which uses multiple non-autoregressive Recurrent Graph Convolutional Networks (RGCN);  3) we validate the superiority of the proposed approach on publicly available medical dataset and evaluate the effect of autoregressive LSTMs on MGMC architectures.


%

%
\section{MATERIALS AND PREPROCESSING}
\label{sec:materialsAndPreprocessing}
%


\label{sec:methods}
\subsection{Dataset and Preprocessing}
We used two publicly available datasets in this work: The Alzheimer's Disease Prediction Of Longitudinal Evolution (TADPOLE) \cite{DATASET:marinescu2018tadpole} obtained from the Alzheimer's Disease Neuroimaging Initiative (ADNI) database (\url{adni.loni.usc.edu}) and the Parkinson's Progressive Marker Initiative (PPMI) dataset \cite{DATASET:marek2011parkinson}.
TADPOLE requires classification of subjects into three categories, normal control (NC), mild cognitive impairement (MCI), and Alzheimer's disease (AD). PPMI requires detection of Parkinson's disease (PD) vs. normal controls (NC). 

In TADPOLE, we used 813 subjects coming from the ADNI protocol with 229 NC, 396 MCI and 188 AD diagnosed at baseline. This dataset contains pre-processed features \cite{DATASET:marinescu2018tadpole} from cerebro-spinal fluid (CSF) markers, magnetic resonance imaging (MRI), positron emission tomography FDG (PET), diffusion tensor imaging (DTI), cognitive assessment scores, genetic information such as alipoprotein E4 (APOE4), and demographic information. 
Further pre-processing entailed a normalization of real-valued TADPOLE features to zero-mean and unit-variance. To match the classification task, we selected only features at baseline, and excluded features containing longitudinal information. We further removed features that were available for less than 10\% of the available entries. In the end, the feature matrix had a dimensionality of 813 $\times$ 435, excluding label information. 

In the PPMI dataset, we used all 75 healthy controls (HC) and 249 subjects with PD. PPMI data consists of brain MRI as well as non-imaging information such as Unified Parkinson's Disease Rating Scale (UPDRS), Montreal Cognitive Assessment (MoCA) scores, and demographic information (age and gender). The MRI information is used as input to the network while non-imaging information is used for the graph construction. As described in our previous GCN CADx approach \cite{MICCAI2019:Kazi}, we pre-processed MRI volumes by co-registering each images to a normative space (SRI24 atlas \cite{atlas_rohlfing2010sri24}) to reduce variability in appearance , and further performed skull stripping using ROBEX \cite{iglesias2011robust}. Then we scaled each volume to an intensity range of [0,1]. Finally, to obtain a lower dimensional representation as input to the graph network, we used encoded raw image intensities coming from a 3D-autoencoder, which was pretrained for towards anomaly detection. We refer the reader to \cite{Preprocessing:baur2018deep} for a detailed discussion on the implementation of the pre-processing and 3D-autoencoder. The output at the bottleneck layer of the 3D-autoecoder was then used as the feature representation of the brain MRI volume.

Notably, our pre-processed PPMI dataset was 100\% feature complete. In contrast, the TADPOLE dataset is inherently incomplete in native form, and was 83\% feature-complete after our pre-processing pipeline. In the experimental section, we further removed known features artificially, to test classification and imputation robustness at various levels of data missingness. For better clarity throughout the rest of the paper, when denoting e.g. 50\% data availability, we refer to the amount of data available at baseline (e.g. 50\% for PPMI, and 41.5\% for TADPOLE). 

\section{METHODS}
\label{sec:methods}
%

        

We first introduce the notation used throughout the rest of the paper in Table \ref{table:notations} then elaborate on key background information in order to provide more context on our proposed approach. 

\begin{table}
\centering
\caption{Description of Notations}
\label{table:notations}
\setlength{\tabcolsep}{3pt}
\begin{tabular}{p{28pt}|p{37pt}|p{152pt}}
\hline
Notation& 
Dimension& 
Description \\
\hline
$\mathbf{X}$ & $n \times m$ & Observed feature matrix with $n$ samples and $m$ features \\
$\mathbf{Y}$ & $n \times c$ & Class label matrix with $n$ samples and $c$ number of class \\
$\mathbf{Z}$ & $n \times (m+c)$ & Concatenated $\mathbf{X}$ and $\mathbf{Y}$ matrices \\
$\mathbf{\hat{X}}$ & $n \times m$ & Predicted feature matrix $\mathbf{X}$  \\
$\mathbf{\hat{Z}}$ & $n \times (m+c)$ & Predicted matrix $\mathbf{Z}$ \\
$\mathbf{\bar{Z}}$ & $n \times (m+c)$ & Predicted matrix $\mathbf{Z}$ from a single RGCN\\

\hline
$||.||_{\operatorname{F}}^2$ & -- & Frobenius norm \\
$||.||_{\operatorname{D,r}}^2$ & -- & Dirichlet norm on the row graph \\
$\mathcal{L}_{ce}(.)$ & -- & Cross-entropy loss \\
$\mathcal{L}_{R}(.)$ & -- & Reconstruction loss from GMC \\
$M^{(i)}$ & -- & The $i$-th meta-information \\
$\mathbf{M}$ & -- & Set containing $\{M^{(1)}, ..., M^{(I)}\}$ \\
$G_{i}$ & -- & The $i$-th graph constructed using meta-information $M^{(i)}$ \\
$\mathbf{\Omega}_{x}$, $\mathbf{\Omega}_{y}$ & -- & Denote whether input features and class lables, respectively, are known (1) or missing (0) \\
$\Theta, \delta$ & -- & Parameters from GCN and LSTM, respectively \\
$\gamma_{\{a,b,c\}}$ & -- & Hyper-parameters weighting loss terms \\
\hline
\end{tabular}
\end{table}
\subsection{Graph Construction}
We use meta-information to construct separate graphs for each dataset. In the TADPOLE dataset, we use meta-information such as age, gender, and genetic risk factor (APOE4), all of which are known risk factors related to AD. For every given meta-information we calculate a separate graph using a pairwise similarity function. An edge between nodes $i$ and $j$ is defined using $W(i,j) = f(M(i), M(j))$ where
\begin{equation}
\label{eq:similarity_function}
f(M(i), M(j)) = 
    \begin{cases}
    1 & \text{if } |(M(i) - M(j)| \leq \theta\\
    0 & \text{otherwise}
    \end{cases}
\end{equation}

$M(i)$ and  $M(j)$ denote meta information of node $i$ and $j$, and $\theta$ denotes a threshold value which is chosen empirically. 

To construct the graphs for the PPMI dataset, we use the same formulation in equation \eqref{eq:similarity_function} and build graphs for every meta-information. Here we again use age and gender, along with two PD-related clinical scores of motor function (UPDRS) and of cognitive function (MoCA) to build the graph, following \cite{MICCAI2019:Kazi}. 

\subsection{Geometric Matrix Completion}

Consider an incomplete feature matrix $\mathbf{X} \in \mathbb{R}^{n\times m}$ where a certain proportion of values is missing at random. The goal is to recover the missing values in this matrix. One solution to this problem is by using rank minimization. However, as this is known to be computationally intractable, an alternative approximation is to constrain the predicted values to be smooth with respect to some geometric structure \cite{MC:Rao15,GMC:Kalofolias14, RGCN:Monti17}. Here a graph structure is built based on the rows or columns of the matrix. Monti et al. \cite{RGCN:Monti17} proposed to solve this using geometric deep learning on graphs, through a combination of GCN and LSTM networks. 
Compared to GMC recommender systems in \cite{RGCN:Monti17}, our CADx problem does not allow us to build a semantically meaningful column graph, especially since features stem from different modalities. Therefore, we modify the GMC  approach to consider only a row graph derived from from patient similarities to model the population. Similarities are computed from meta-features using the metric in equation \ref{eq:similarity_function}. Pair-wise similarities between nodes in the population graph connect patients that share the same risk-factor characteristics. The row graph is then represented as $G = (V, E, \mathbf{W})$, with vertices $V = \{1,2,...,n\}$, and edges $E \subseteq V \times V$, which are weighted with non-negative weights. We represent the graph with a symmetric adjacency matrix $\mathbf{W} \in \mathbb{R}^{n\times n}$. The geometric matrix completion problem when considering only the row graph reduces to solving:

\begin{equation}
\label{eq:gmc}
    \ell(\Theta, \delta) =
	 ||\mathbf{\hat{X}}_{\Theta,\delta}||_{D}^2 +
	 \frac{\gamma}{2} ||\mathbf{\Omega}_x \circ (\mathbf{\hat{X}}_{\Theta,\delta} - \mathbf{X})||_{\operatorname{F}}^2
\end{equation}


where $\mathbf{\hat{X}_{\Theta,\delta}}$ is the predicted matrix conditioned on the parameters of the GCN and LSTM. In equation \eqref{eq:gmc}, the first term on the right is equal to $\operatorname{tr}(\mathbf{\hat{X}}^T \mathbf{L} \mathbf{\hat{X}})$ which contains a rescaled graph Laplacian ($\mathbf{L} \in \mathbb{R}^{n \times n}$) term such that its eigenvalues are in the interval $[-1, 1]$. This term keeps the prediction smooth with respect to the row graph structure. 




GMC can also be extended to multi-target training on heterogeneous matrix entries. Consider a matrix $\mathbf{Z} \in \mathbb{R}^{n \times (m + c)}$, which contains a mixture of feature and label information, which is implemented by concatenation of the feature matrix $\mathbf{X} \in \mathbb{R}^{n \times m}$ and class label matrix $\mathbf{Y} \in \mathbb{R}^{n \times c}$, similarly to Goldberg et al. \cite{MC:Goldberg10}. Following equation \eqref{eq:gmc}, we can add a classification loss term on the imputed class label matrix \cite{Vivar2018}. The combined loss for completion of matrix $\mathbf{Z}$ is then:
\begin{multline}
\label{eq:mGMC}
\ell(\Theta, \delta) =
 \frac{\gamma_a}{2} ||\mathbf{\hat{Z}}_{\Theta, \delta}||_{D}^2
 + \frac{\gamma_b}{2} ||\mathbf{\Omega}_{x} \circ (\mathbf{\hat{Z}}_{\Theta, \delta} - \mathbf{Z})||_{\operatorname{F}}^2\\
  + \gamma_c(\mathcal{L}_{ce}(\mathbf{\hat{Z}}_{\Theta, \delta} \circ \mathbf{\Omega}_{y}, \mathbf{Z} \circ \mathbf{\Omega}_{y}))
 \end{multline}

where $\mathbf{\hat{Z}_{\Theta,\delta}}$ is the predicted matrix containing predictions for both $\mathbf{\hat{X}}$ and $\mathbf{\hat{Y}}$.


%
\subsection{Multigraph Geometric Matrix Completion}
\label{sec:Methods:MGMC}
%
MGMC\footnote{Code: \url{https://github.com/pydsgz/MGMC}} consists of multiple non-autoregressive RGCNs and Transformer-like self-attention. We first describe the motivation why we use multiple RGCNs then elaborate on the self-attention inspired aggregation scheme including the use of non-autoregressive RGCNs. First, as we described in our previous works \cite{ISBI:Kazi19SelfAtt, MICCAI2019:Kazi}, the rules for constructing a population graph from a medical dataset are crucial to the accuracy of a GCN's downstream task, e.g. diagnostic classification accuracy. Instead of collapsing all meta-features into a single patient similarity measure, we therefore construct multiple graphs, one for each meta-feature. We then propose to integrate multi-graph GCNs into matrix completion by training a dedicated GCN and LSTM for each graph in an end-to-end manner. We do this to learn better imputed feature representations for each graph which could be useful in the downstream classification task.

\begin{figure}[]
  \centering
    \includegraphics[width=.5\textwidth]{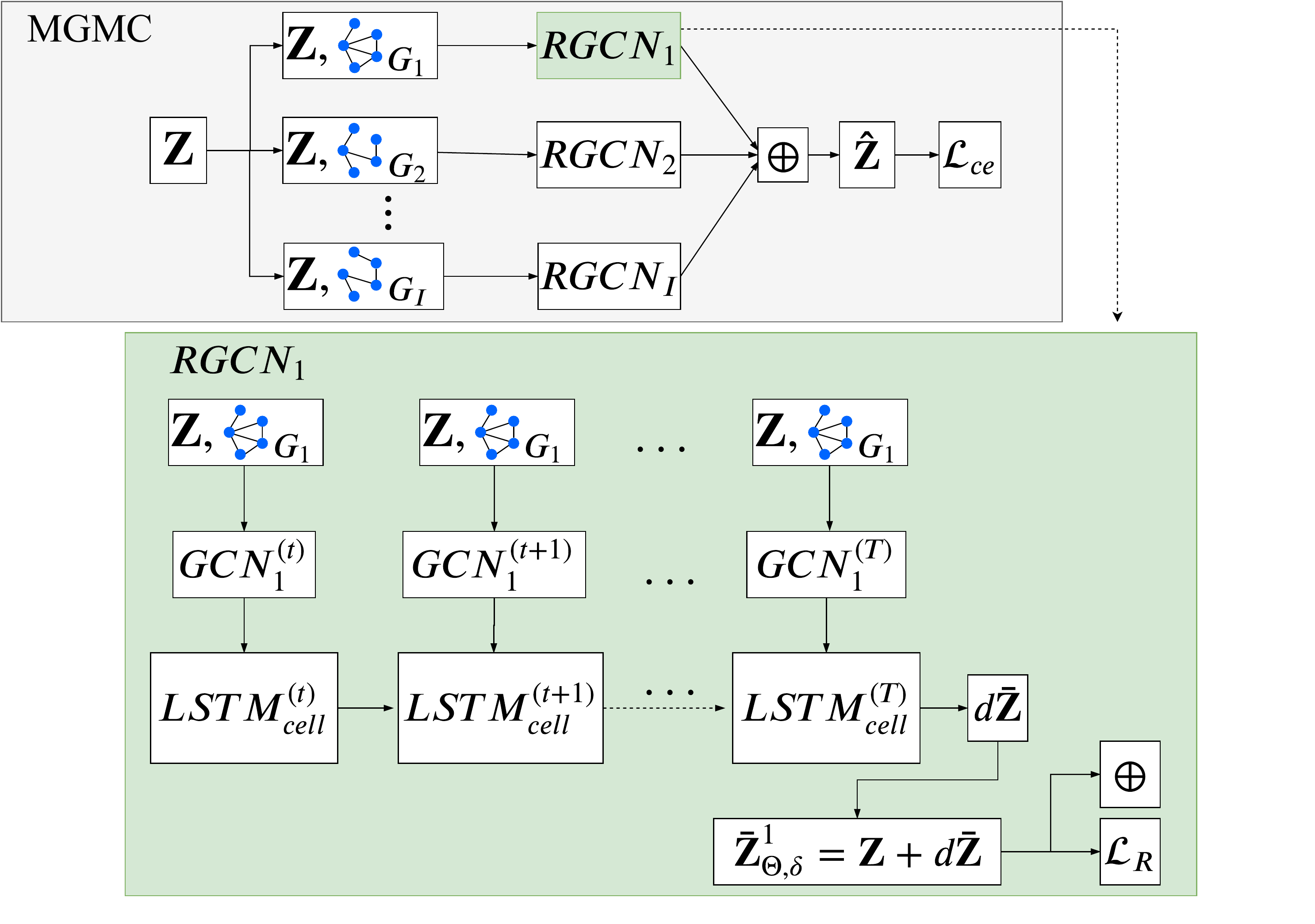}
  \caption{Network architecture of MGMC which uses multiple Recurrent Graph Convolutional Network (RGCN) (top) including non-autoregressive RGCN layer (bottom).}
  \label{fig:mrgcn_architecture}
\end{figure}

To aggregate separate signals from parallel RGCNs, we use a self-attention aggregation mechanism inspired by Transformer networks \cite{Vaswani2017}. We do this by training separate RGCNs (which consists of GCN and LSTM) in an end-to-end manner as shown in figure \ref{fig:mrgcn_architecture}, then aggregate graph outputs using the weights learned from the self-attention layer. Furthermore, we use muliple RGCNs, wherein each (unrolled) RGCN consists of a GCN and a non-autoregressive LSTM. Although the use of multiple graphs and LSTMs have been used in previous methods (\cite{RGCN:Monti17, Coates2019}, one important difference of our proposed approach is the use of non-autoregressive LSTMs. As shown in figure \ref{fig:mrgcn_architecture}, we only use the original input feature as input to the next timestep including the learned parameters from the previous LSTM cell-block. Such a non-autoregressive strategy is motivated in several ways. First, it limits the number of neighborhood hops and graph signal diffusion steps, as the input feature matrix to the GCN layer is the same at every time-step in the RGCN. Second, it allows the model to have better control on which graph-relevant information is useful for the imputation and downstream classification task. Third, by using the original input features as prior information at every optimization step, we reinforce the reconstruction of the input data, and prevent the model from diverging from the input data. As a result, this strategy prevents the model from suggesting non-realistic features as outputs. For the GCN layers, we use a Cheb-Net implementation \cite{GCN:Defferrard16, RGCN:Monti17}. This uses a Chebyshev polynomial basis ($\sum^{K}_{k=0} T_k (\Tilde{L}) X \Theta_{k}$) to represent the spectral filters. For a more in-depth discussion regarding deep learning on graphs we refer the reader to \cite{GCN:Bronstein17}. The optimization loss for multi-graph GMC then boils down to solving:
\begin{multline}
\label{eq:multiGMC}
\ell(\Theta, \delta) =
\sum_{i}^{M}
 (\frac{\gamma_a}{2} ||\mathbf{\bar{Z}}^{(i)}_{\Theta,\delta}||_{D,r}^2
 + \frac{\gamma_b}{2} ||\mathbf{\Omega}_{x} \circ (\mathbf{\bar{Z}}^{(i)}_{\Theta,\delta} - \mathbf{Z})||_{\operatorname{F}}^2)
 \\ + \gamma_c(\mathcal{L}_{ce}(\mathbf{\hat{Z}}_{\Theta,\delta} \circ \mathbf{\Omega}_{y}, \mathbf{Z} \circ \mathbf{\Omega}_{y}))
 \end{multline}

where $\mathbf{\bar{Z}}^{(i)}_{\Theta,\delta}$ is the i-th predicted matrix based from the i-th graph (noting that this is conditioned on the parameters of the i-th GCN and LSTM) and $\mathbf{\hat{Z}}^{(i)}_{\Theta,\delta}$ is the aggregated predicted matrix coming from all GCNs and LSTMs.

\section{Results}
\label{sec:results}
%

\subsection{Implementation Details.}
We split the dataset into 10\% test and 90\% train (of which 10\% as validation set) on all methods. For all deep learning based methods we use Adam optimization \cite{kingma2014adam}, with implementations in PyTorch \cite{paszke2019pytorch}, on a workstation with single-GPU (Nvidia GTX 1080 Ti). We automatically determine hyperparameters in equation \eqref{eq:multiGMC} using hyperparameter optimization on the validation set with 120 iterations \cite{bergstra2015hyperopt}, with the following search spaces for the Chebyshev Polynomial parameters ($K \in  \text{range}(1, 20)$), learning\_rate = uniform($[0.00001, 0.1]$), intermediate\_layer\_hidden\_units $\in \text{range}(8, 512)$, and $\gamma_{(a, b, c)}$  = uniform($[0.001, 1000$])). 

We compared the proposed method with shallow learning methods in machine learning and state-of-the-art graph-based methods which have shown to be highly effective for disease prediction. For shallow learning, we used  Logistic Regression (LR) as the linear baseline, and Random Forest (RF) \cite{MODEL:breiman2001randomforest} as a competitive non-linear baseline. Previous graph-based methods included GCNs and GMC. As several algorithms (LR, RF and GCN) assume feature-completeness, we first need to impute the missing values in the feature matrix. We used three approaches to accomplish this: the commonly used mean imputation method, kNN imputation \cite{troyanskaya2001KNN_imputation}, and the state-of-the-art MICE algorithm, the latter with random regression forests for estimation \cite{buuren2010mice}. To test imputation performance, we artificially reduce the percentage of known data in the ADNI/PPMI feature matrices and perform imputation/classification at \{100,75,50,25\}\% data availability.  We use Scikit-learn \cite{pedregosa2011scikit} implementations for cross-validation, pre-processing, imputation and shallow classifier models (LR and RF).


To make baseline algorithms as competitive as possible, we also perform hyperparameter optimization (also 120 max. iterations) for the standard machine learning models (i.e. LR and RF) \cite{komer2014hyperopt-sklearn}. We concatenate the meta-features (e.g. demographics) with the feature vectors for all baseline methods, to further ensure fairness, as our proposed graph-based method utilizes this information as well (for graph construction). 

\subsection{PPMI and TADPOLE Dataset Results.} 
In Fig. \ref{fig:ppmi_tp_results} (left panel), we plot classification and imputation results on the PPMI dataset. It is clearly visible that the classification accuracy and AUC of LR and RF models are consistently lower than our proposed approach. Since both models require feature completeness, they operate on imputed features. In the same AUC boxplots, we observe that the proposed approach has significantly higher AUC than standard machine learning models and previous graph-based methods (GCN and GMC). We can also observe that the imputed values of the proposed approach have a much lower absolute deviation (RMSE) from the real values than mean, MICE, kNN, and GMC imputation. 

\begingroup
\setlength{\tabcolsep}{.5pt} 
\renewcommand{\arraystretch}{.1} 
\begin{figure*}
    \begin{tabular}{cc}
 PPMI & TADPOLE \\
  \includegraphics[clip, trim=0 0 0 0, width=.49\textwidth]{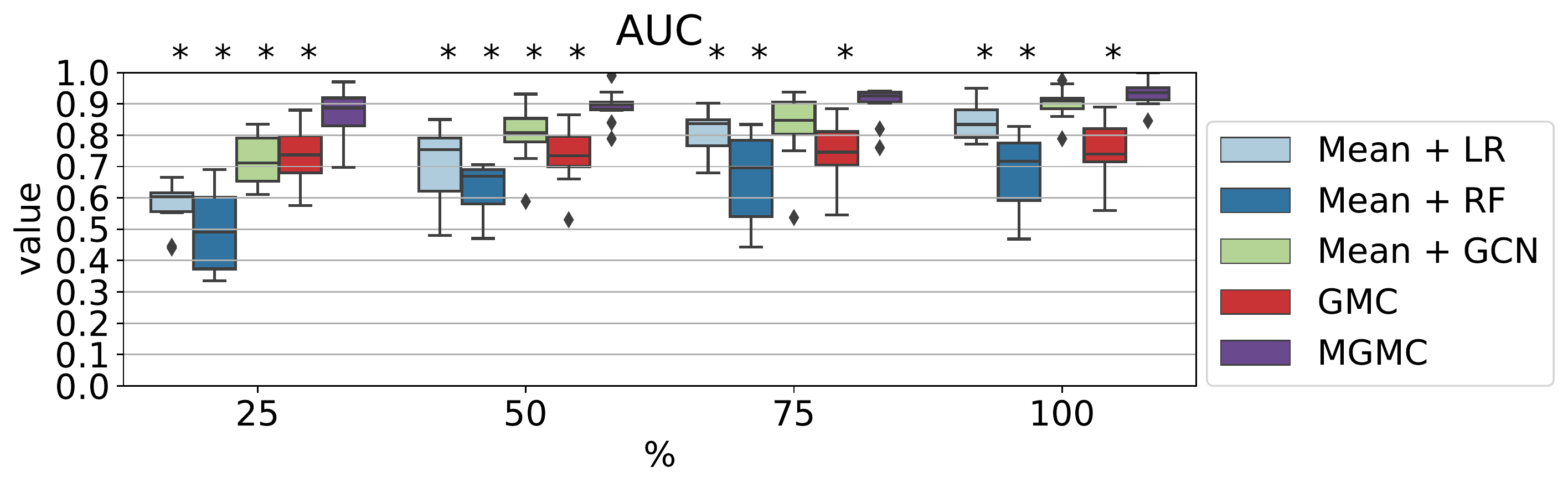} 
  &
  \includegraphics[clip, trim=0 0 0 0, width=.49\textwidth]{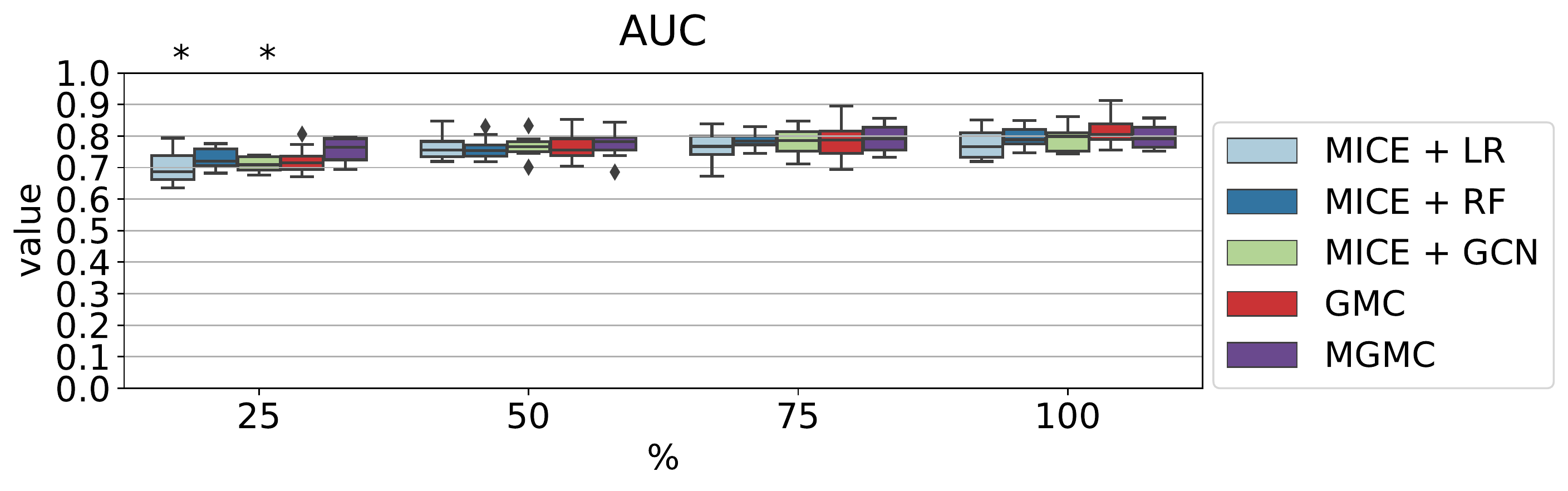}
  \\
  \includegraphics[clip, trim=0 0 0 0, width=.49\textwidth]{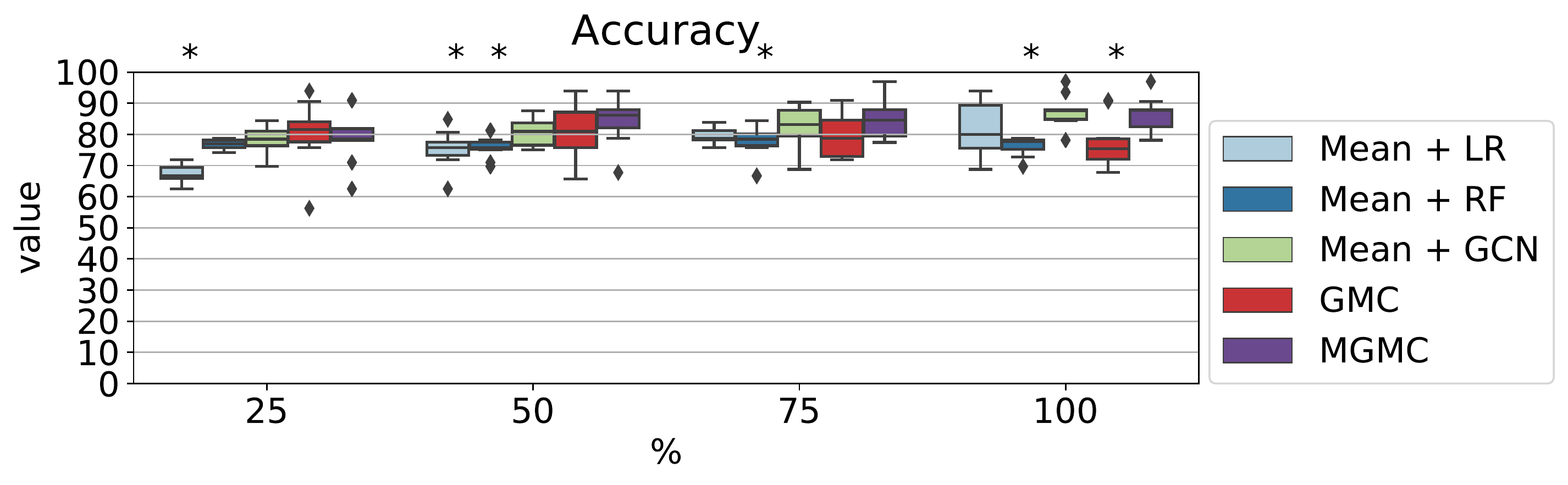}
  &
  \includegraphics[clip, trim=0 0 0 0, width=.49\textwidth]{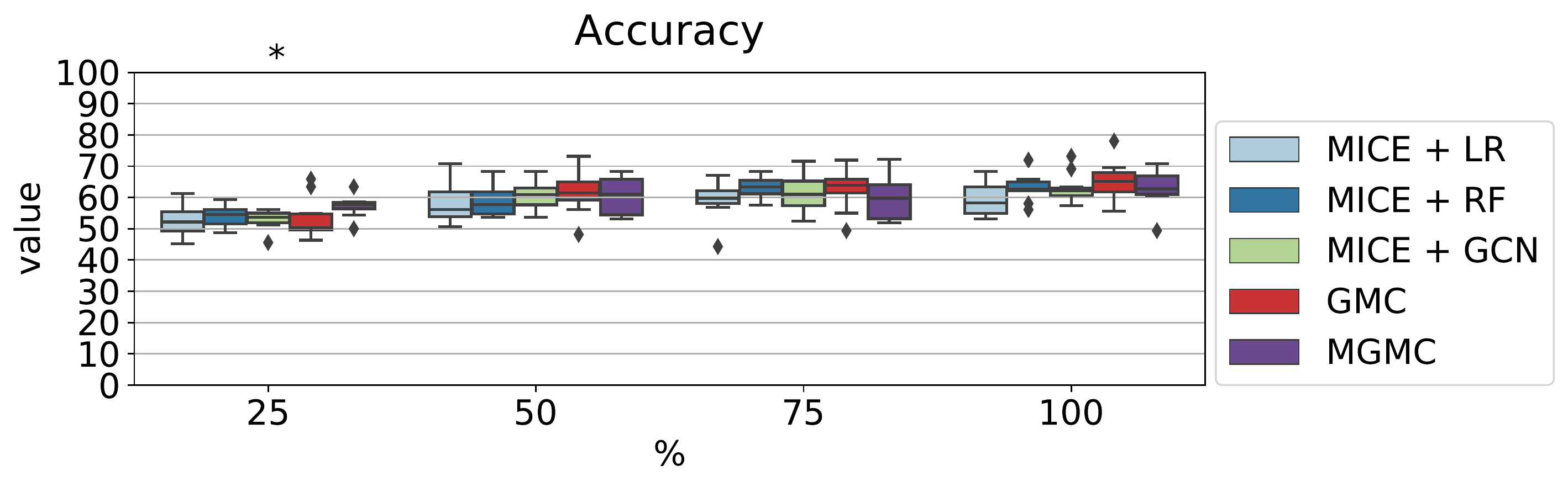}
  \\
  \includegraphics[width=.47\textwidth]{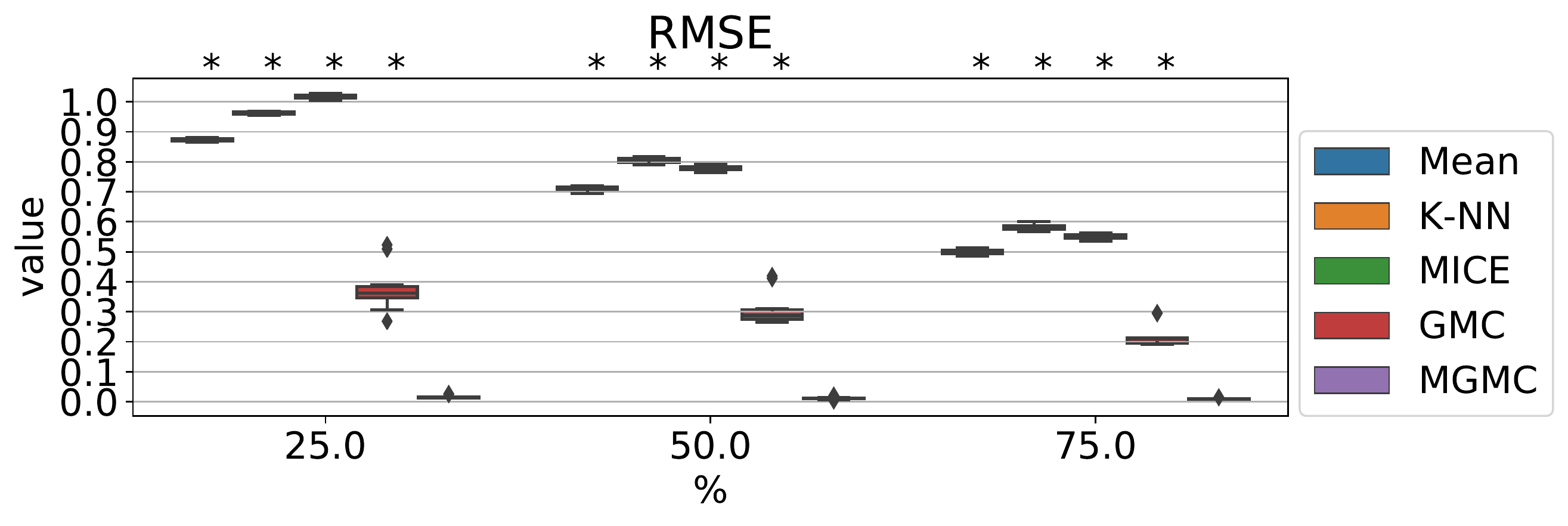}
  &
  \includegraphics[width=.47\textwidth]{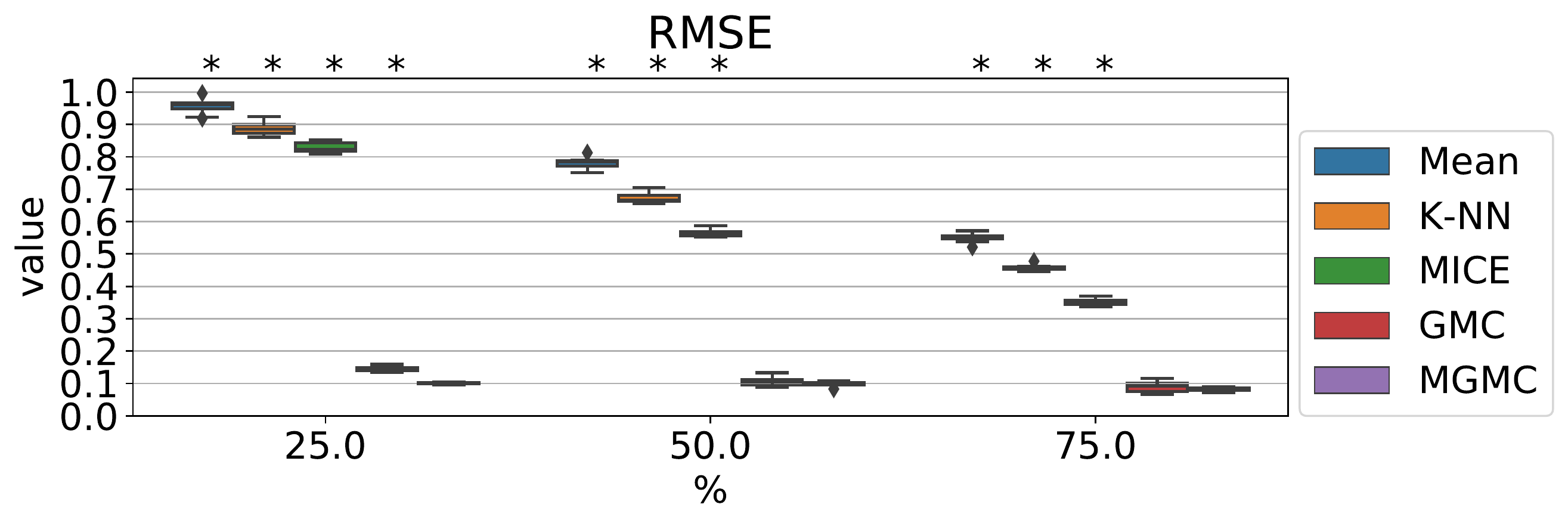}
  \end{tabular}

  \caption{PPMI (left) and TADPPOLE (right) results: ROC-AUC (top), Accuracy (middle) and RMSE (bottom). Asterisk symbols (*) denote that the tested model is statistically significantly different (two-tailed Wilcoxon rank-sum test, $p \leq 0.05$) to our proposed model (MGMC). X-axis values denote the percentage of available/known features prior to imputation and model training.}
  \label{fig:ppmi_tp_results}
\end{figure*}
\endgroup
Further, in Fig. \ref{fig:ppmi_tp_results} (right panel), compared to PPMI, the classification accuracy does not benefit as clearly from the population graph or imputation in our method. MultiGMC still classifies better than linear methods, with a significant advantage at lower percentage of data availability (25\%). As in PPMI, however, the imputation results in terms of RMSE are significantly and consistently better than the baseline imputation methods (mean, MICE, kNN). Similar to PPMI, the trend is visible that baseline imputation methods impute with higher RMSE errors as fewer data is available in the feature matrix, whereas the MultiGMC provides fairly robust imputation results.


We performed ablative experiments to see how non-autoregressive LSTMs affect the imputation and classification performance. In Fig. \ref{fig:ppmi_tp_abla} left, we observe that for the PPMI dataset, the non-autoregressive model yields significantly better results in terms of AUC and Accuracy at all levels of data missingness. For the TADPOLE dataset (Fig. \ref{fig:ppmi_tp_abla} right), the proposed method classifies comparably well at 50\%, 75\% and 100\% data availability, but significantly outperforms the auto-regressive model at 25\% data availability, demonstrating better classification robustness at lower levels of data availability. 

\begingroup
\setlength{\tabcolsep}{.5pt} 
\renewcommand{\arraystretch}{.1} 
\begin{figure*}
  \begin{tabular}{cc}
  PPMI & TADPOLE \\
  \includegraphics[width=.5\textwidth]{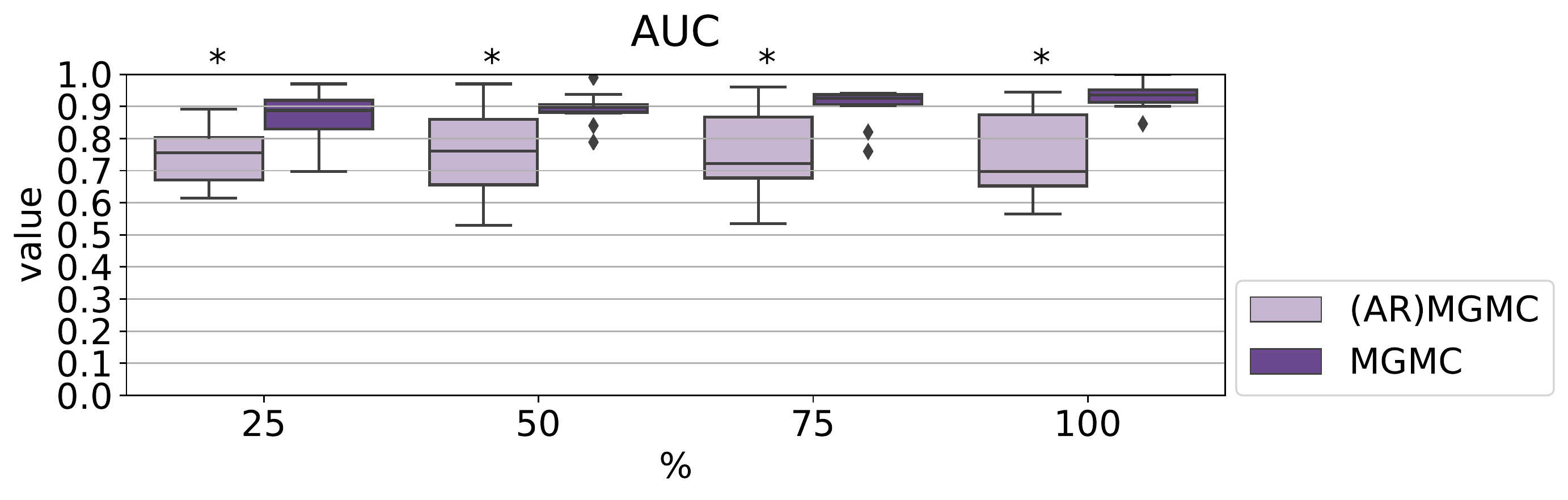} 
  &
  \includegraphics[width=.5\textwidth]{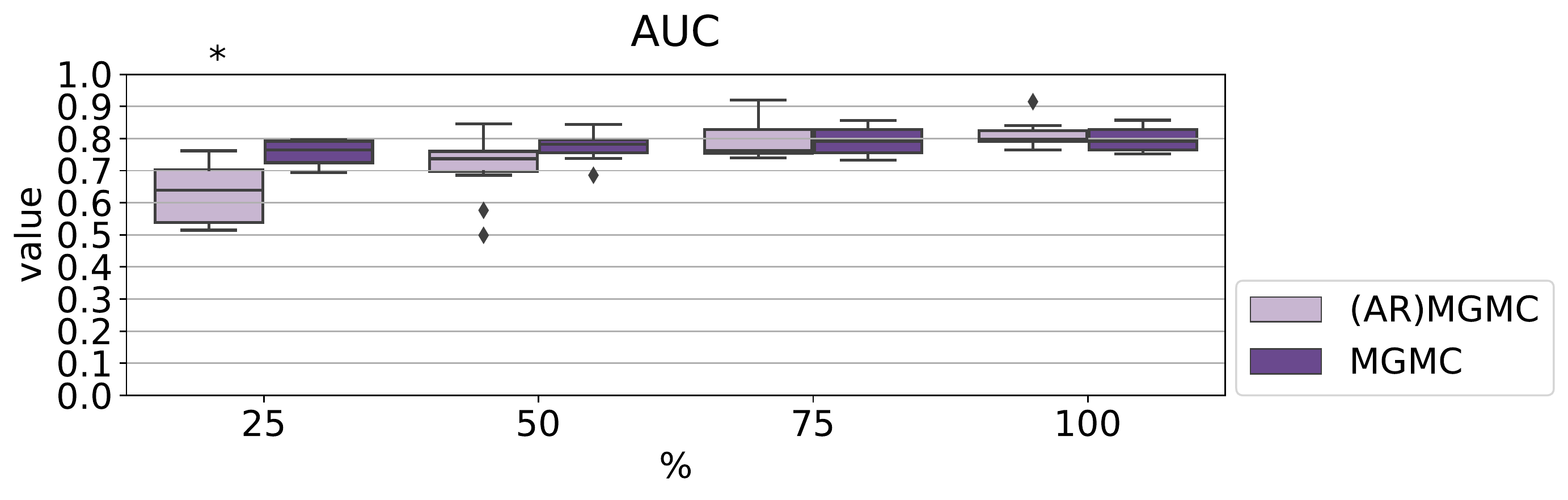}
  
  \\
  \includegraphics[clip, trim=0 0 0 0, width=.5\textwidth]{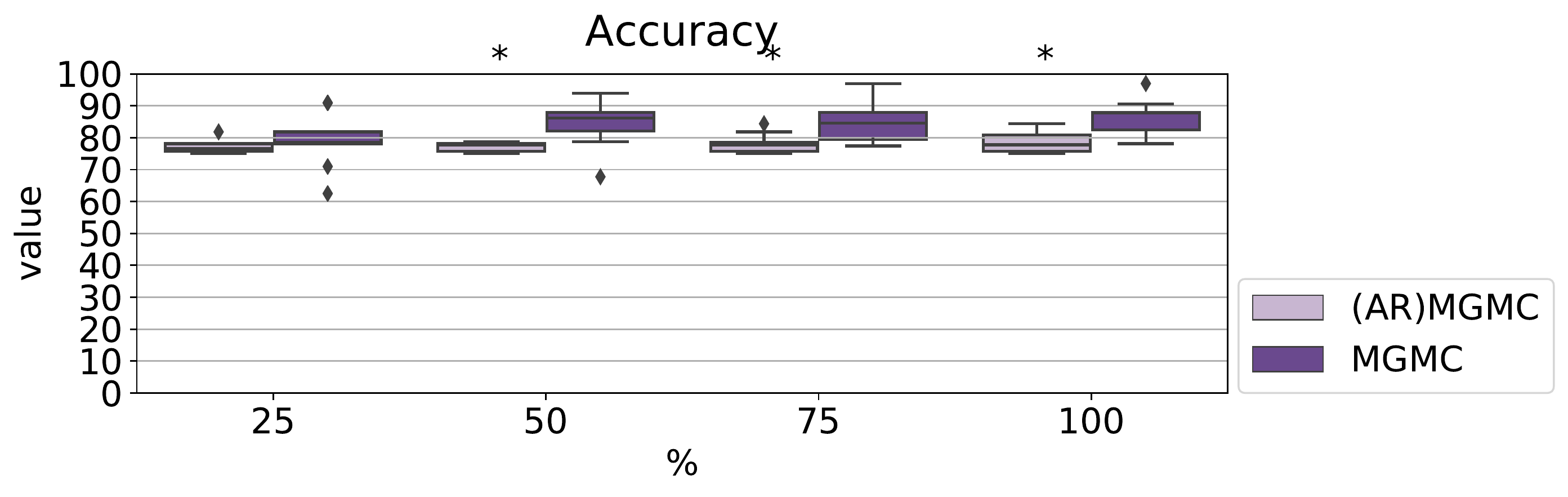}
  &
  \includegraphics[clip, trim=0 0 0 0, width=.5\textwidth]{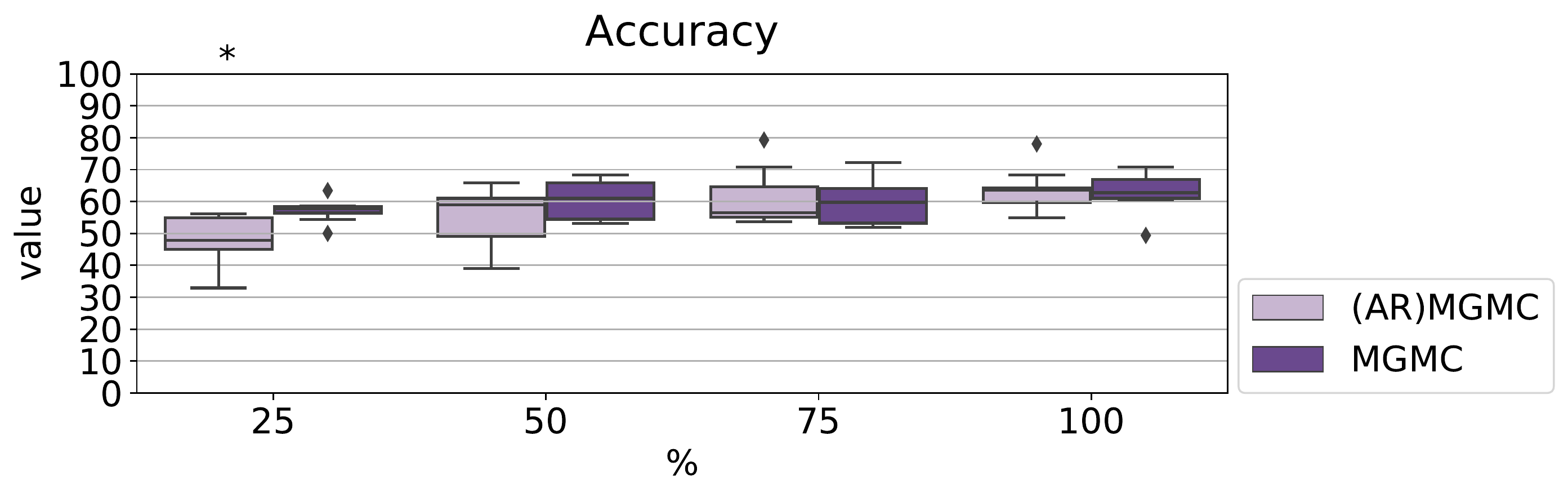}
  \end{tabular}
  \caption{ PPMI (left) and TADPOLE (right) ablation results. ROC-AUC (top) and accuracy (bottom) results on test dataset. (*) denote model is statistically significantly different ($ p \leq  0.05$) to proposed model. Values in x-axis denote relative percentage of features which are available to the network.}
  \label{fig:ppmi_tp_abla}
\end{figure*}
\endgroup

%
\section{Discussion}
\label{sec:discussion}
%

\subsection{Classification performance when using all available data}
In PPMI, we observe that our proposed approach achieves significantly better classification in terms of AUC and Accuracy for PD prediction when compared with standard ML models (LR and RF), as shown in Fig. \ref{fig:ppmi_tp_results} left. In TADPOLE, we observe that our approach is mostly on a par with baseline ML methods and SOTA approaches from literature. As mentioned in the dataset descriptions, PPMI is 100\% feature-complete at baseline, whereas TADPOLE is only 83\% complete at baseline. It is noteworthy that at 100\% data availability, MGMC already performs imputation in TADPOLE, but we cannot validate the imputed values due to a lack of groundtruth data for those missing features. Compared to previous studies, Zhou et al. \cite{TMI2019:Zhou2019} reported $\sim$60\% classification accuracy and $\sim$0.6 AUC for the same AD classification problem posed in this paper for the TADPOLE dataset. Gray et al. \cite{NeuroImage2013:RandomForest_Gray} reported $\sim$60\% classification accuracy and $\sim$0.7 AUC. In our study, we also achieve a classification accuracy on the order of $\sim$60\%, however with higher AUC values on the order of $\sim$0.8. To interpret these results, we recall that the accuracy metric represents the number of true positive and true negative cases among the total population, at a fixed threshold of the model's posterior. In comparison, the ROC-AUC gives an estimate of the likelihood that a classifier simultaneously achieves a high true positive rate and low false positive rate. This indicates that MGMC, compared to related works, and compared to baseline models at 25\% data availability, achieves a more robust classification outcome, not only in terms of sensitivity, but also in form of a lower likelihood for type I errors. A likely reason for the AUC difference of $\sim$0.1 compared to \cite{NeuroImage2013:RandomForest_Gray} is that earlier (2013) versions of the ADNI dataset had a smaller sample size, which also makes comparisons to our work somewhat unfair. Compared to \cite{TMI2019:Zhou2019}, the AUC difference of $\sim$0.2 can be likely attributed to the use of multi-graph convolutions in our work, which are trained end-to-end in a semi-supervised manner.

\subsection{Classification performance with artificially removed data}
To investigate the robustness of MGMC and baseline methods with respect to missing data, we randomly reduce the amount of available data in the feature matrix relative to the number of observed entries at baseline, as shown in Fig. \ref{fig:ppmi_tp_results}. We observe that the proposed approach has better and more stable classification and imputation results for PD prediction in PPMI when more information is missing. This effect is particularly visible in the AUC values, which may increase in standard deviation over the ten cross validation folds, but stay relatively stable in terms of median values above 0.9, even at low level of data availability around 25\%. In comparison, LR, RF and GCN suffer from a noticeable drop in classification robustness. Interestingly, the single-graph GMC also yields relatively constant AUC values, but at a significantly lower level than MGMC. This has two important implications. First, end-to-end learning of simultaneous imputation and classification, e.g. via geometric matrix completion, noticeably improves the robustness of the CADx model towards the level of incompleteness in datasets. Second, multiple graphs in parallel, e.g. fused by self-attention, improve both downstream tasks significantly, compared to using a single-graph. It is important to note, however, that the two downstream tasks do not always benefit equally. In TADPOLE, for example,  we observe a comparably stable classification performance at 75\%, 50\% and 25\% data availability. However, a similar behaviour is observed for all other classifiers, and all classifiers in general classify similarly well. Only at 25\% data availability, MGMC significantly outperforms GCN and LR (the latter two with MICE imputation), but we consider this a negligible advantage. Clearly, the main benefit of the proposed method on TADPOLE data lies not in an improved classification, but in a significantly more accurate imputation of missing values.

\subsection{Joint classification and imputation performance}
Most related literature in CADx naturally puts a focus on classification performance. Imputation is an often overlooked factor, even though it plays an important role in population-based and multi-modal studies in medicine, as data missingness is a common problem here \cite{IMPUTATION:Little12}. When looking at imputation performance in Fig. \ref{fig:ppmi_tp_results}, we can see that the proposed approach is able to significantly outperform standard imputations such as mean, kNN, and MICE at all levels of missingness, on both datasets. This suggests that the proposed approach is able to take advantage of using known (semi-supervised) class label information in order to impute the features while simultaneously predicting the unknown class labels. It further suggests that the proposed method learns more class relevant feature representations compared to standard imputation approaches (mean, kNN, and MICE). We can also observe that population modeling and graph incorporation cannot properly compensate for sub-optimal imputation, as Mean+GCN results in PPMI are all inferior to the matrix completion methods. Unfortunately, we cannot compare the imputation results with other matrix completion works in literature \cite{SWMC:Thung16, MedIA2018:Thung2018, TMI2019:Zhou2019, Coates2019}, as those works do not report imputation fidelity, e.g. via RMSE. However, we can compare classification performance on TADPOLE data with \cite{TMI2019:Zhou2019}, who used the same subjects (examinations at baseline) and classes (NC, MCI and AD) in TADPOLE as we did in our study. Here, authors explored classification performance of their proposed method, given 10\% and 20\% data missingness on either the MRI or SNP modality. As authors in \cite{TMI2019:Zhou2019} report, the results of our proposed approach are in line with their classification accuracy results at 20\% data missingness (~60\% accuracy) which corroborates our results on 75\% data availability in Fig. \ref{fig:ppmi_tp_results} right. Finally, it is noteworthy that our proposed approach achieved a much better and stable classification performance for the PPMI prediction task than for the TADPOLE prediction task. A possible explanation is that distinguishing healthy controls from PD may be a much simpler classification task than the three-class classification problem in TADPOLE (NC vs. MCI vs. AD). This notion is supported by clinical studies arguing that distinguishing NC, MCI, and AD based on clinical characteristics is a difficult problem at baseline \cite{langa2014MCIADbaseline}.



\subsection{Ablation experiments}
In section \ref{sec:Methods:MGMC}, we described our proposed improvements for usage of multiple RGCNs, specifically the usage of non-autoregressive LSTMs over autoregressive ones. Autoregressive RGCNs always use the output from the previous timestep and information from the previous LSTM cell-block as input. In contrast, non-autoregressive RGCNs always use the original input features as input at every timestep. Our motivation for using non-autoregressive LSTMs in MGMC is that the current output is always conditioned on the original input features. Intuitively,  this should help the reconstructed output to avoid diverging from the input data, which is a desirable behaviour in matrix completion. Here, we perform and discuss an ablation experiment, where we compare the effect of both, as shown for PPMI and TADPOLE in Fig. \ref{fig:ppmi_tp_abla}.  We observe that by using non-autoregressive LSTMs, we obtain a significantly better classification performance for all levels of data availability in PPMI. In TADPOLE, this tendency is not as clear, and a significant improvement is only achieved at 25\% relative available data. At 50\%, 75\% and 100\% available data, non-autoregressive LSTMs do not improve classification, but neither do they worsen the performance. This result suggests that it is indeed preferable to use non-autoregressive LSTMs in each parallel graph branch in MGMC. We attribute this to the intuitive notion explained above: by conditioning the reconstructed output on the original input data at every optimization timestep, we stabilize the reconstruction and achieve a better classification performance.



\subsection{Overall implications}


The main differences of our proposed approach to recent works that use RGCNs for matrix completion \cite{Vivar2018, Coates2019, RGCN:Monti17} are three-fold, namely i) the use of multiple LSTMs which are non-autoregressive, ii) the use of self-attention weighting to aggregate information from iii) multiple graphs representing different neighborhood relationships between patients in the population. Previous RGCN/GMC methods \cite{Vivar2018, Coates2019} use a single LSTM, while in our approach, we utilize one separate LSTM for every graph, which results to multiple recurrent graph convolutional networks (MRGCN). Notably, Monti et al. \cite{RGCN:Monti17} also use a multi-graph formulation, but their approach differs from our method, since they consider both the rows and columns of the feature matrix as two separate graph structures. Instead, in this work, we consider multiple meta-information as separate graphs that contain rows of a feature matrix as the node features, similarly to \cite{MICCAI2019:Kazi}.


Finally, our work has certain limitations, which may suggest interesting avenues for future contributions. Following \cite{GCN:Parisot17, GCN:Parisot18MedIA}, our graph construction heuristic assumes a simple static graph. An alternative approach would be to use the meta-information and the feature matrix information in parallel to build or learn the graph adjacency. 
Both approaches could potentially lead to better performances of the downstream tasks (classification and imputation). 
Another limitation is that we benchmarked our proposed MGMC method to several baseline methods (LR and RF) which are all inductive learning approaches. In contrast, our approach is inherently transductive, as we rely on spectral graph convolutions in the parallel graph-convolutional layers. We believe that it should be possible to incorporate imputation losses into the objective functions of GraphSAGE \cite{hamilton2017graphsage} or GAT \cite{velickovic2018gat} to obtain an inductive form of MGMC, and it is worth investigating whether the same benefits can be observed as in our experiments. Furthermore, future works could compare against other non-deep learning based techniques that tackle missing data such as \cite{wang2016tsvm_based_for_missing_data} and \cite{venugopalan2019clustering_based_for_missing_data}.
\section{Conclusion}
\label{sec:conclusion}
In conclusion, we propose a novel automatic disease classification method which can handle multimodal data with missing information, a common setup in medical population based studies and datasets. We accomplish this by using Multi-graph Geometric Matrix Completion (MGMC). We train our architecture through Multiple Recurrent Graph Convolutional Networks, which are optimized in an end-to-end manner. Experimental results suggest the effectiveness of our proposed approach on two well-known and challenging population based studies of neurodegenerative Parkinson's and Alzheimer's diseases. Furthermore, ablation experiments highlight the importance of using non-autoregressive LSTM including the effect of self-attention weighting. These results could serve as a baseline for future works on disease classification in incomplete datasets. In addition, this could be useful in other domains where incomplete, multimodal, and high-dimensional data is an issue.
\section*{Acknowledgment}
\footnotesize{Data collection and sharing for this project was funded by the Alzheimer's Disease Neuroimaging Initiative (ADNI) (National Institutes of Health Grant U01 AG024904) and DOD ADNI (Department of Defense award number W81XWH-12-2-0012). ADNI is funded by the National Institute on Aging, the National Institute of Biomedical Imaging and Bioengineering, and through generous contributions from the following: AbbVie, Alzheimer’s Association; Alzheimer’s Drug Discovery Foundation; Araclon Biotech; BioClinica, Inc.; Biogen; Bristol-Myers Squibb Company; CereSpir, Inc.; Cogstate; Eisai Inc.; Elan Pharmaceuticals, Inc.; Eli Lilly and Company; EuroImmun; F. Hoffmann-La Roche Ltd and its affiliated company Genentech, Inc.; Fujirebio; GE Healthcare; IXICO Ltd.; Janssen Alzheimer Immunotherapy Research \& Development, LLC.; Johnson \& Johnson Pharmaceutical Research \& Development LLC.; Lumosity; Lundbeck; Merck \& Co., Inc.; Meso Scale Diagnostics, LLC.; NeuroRx Research; Neurotrack Technologies; Novartis Pharmaceuticals Corporation; Pfizer Inc.; Piramal Imaging; Servier; Takeda Pharmaceutical Company; and Transition Therapeutics. The Canadian Institutes of Health Research is providing funds to support ADNI clinical sites in Canada. Private sector contributions are facilitated by the Foundation for the National Institutes of Health (www.fnih.org). The grantee organization is the Northern California Institute for Research and Education, and the study is coordinated by the Alzheimer’s Therapeutic Research Institute at the University of Southern California. ADNI data are disseminated by the Laboratory for Neuro Imaging at the University of Southern California.}

\ifCLASSOPTIONcaptionsoff
  \newpage
\fi
\bibliographystyle{IEEEtran}
\bibliography{bibtex/bib/Bibliography.bib}

\begin{thebibliography}{10}
\providecommand{\url}[1]{#1}
\csname url@samestyle\endcsname
\providecommand{\newblock}{\relax}
\providecommand{\bibinfo}[2]{#2}
\providecommand{\BIBentrySTDinterwordspacing}{\spaceskip=0pt\relax}
\providecommand{\BIBentryALTinterwordstretchfactor}{4}
\providecommand{\BIBentryALTinterwordspacing}{\spaceskip=\fontdimen2\font plus
\BIBentryALTinterwordstretchfactor\fontdimen3\font minus
  \fontdimen4\font\relax}
\providecommand{\BIBforeignlanguage}[2]{{%
\expandafter\ifx\csname l@#1\endcsname\relax
\typeout{** WARNING: IEEEtran.bst: No hyphenation pattern has been}%
\typeout{** loaded for the language `#1'. Using the pattern for}%
\typeout{** the default language instead.}%
\else
\language=\csname l@#1\endcsname
\fi
#2}}
\providecommand{\BIBdecl}{\relax}
\BIBdecl

\bibitem{BHI:Gamache18}
R.~Gamache, H.~Kharrazi, and J.~Weiner, ``Public and {Population} {Health}
  {Informatics}: {The} {Bridging} of {Big} {Data} to {Benefit} {Communities},''
  \emph{Yearbook of Medical Informatics}, vol.~27, no.~01, pp. 199--206, Aug.
  2018.

\bibitem{BHI:Machluf17}
Y.~Machluf, O.~Tal, A.~Navon, and Y.~Chaiter, ``From {Population} {Databases}
  to {Research} and {Informed} {Health} {Decisions} and {Policy},''
  \emph{Frontiers in Public Health}, vol.~5, p. 230, Sep. 2017.

\bibitem{BHI:Dash19}
S.~Dash, S.~K. Shakyawar, M.~Sharma, and S.~Kaushik, ``Big data in healthcare:
  management, analysis and future prospects,'' \emph{Journal of Big Data},
  vol.~6, no.~1, p.~54, Dec. 2019.

\bibitem{IMPUTATION:Little12}
R.~J. Little, R.~D'Agostino, M.~L. Cohen, K.~Dickersin, S.~S. Emerson, J.~T.
  Farrar, C.~Frangakis, J.~W. Hogan, G.~Molenberghs, S.~A. Murphy, J.~D.
  Neaton, A.~Rotnitzky, D.~Scharfstein, W.~J. Shih, J.~P. Siegel, and H.~Stern,
  ``The {Prevention} and {Treatment} of {Missing} {Data} in {Clinical}
  {Trials},'' \emph{New England Journal of Medicine}, vol. 367, no.~14, pp.
  1355--1360, Oct. 2012.

\bibitem{DATASET:marinescu2018tadpole}
R.~V. Marinescu, N.~P. Oxtoby, A.~L. Young, E.~E. Bron, A.~W. Toga, M.~W.
  Weiner, F.~Barkhof, N.~C. Fox, S.~Klein, D.~C. Alexander \emph{et~al.},
  ``Tadpole challenge: Prediction of longitudinal evolution in alzheimer's
  disease,'' \emph{arXiv preprint arXiv:1805.03909}, 2018.

\bibitem{DATASET:marek2011parkinson}
K.~Marek, D.~Jennings, S.~Lasch, A.~Siderowf, C.~Tanner, T.~Simuni, C.~Coffey,
  K.~Kieburtz, E.~Flagg, S.~Chowdhury \emph{et~al.}, ``The parkinson
  progression marker initiative (ppmi),'' \emph{Progress in neurobiology},
  vol.~95, no.~4, pp. 629--635, 2011.

\bibitem{AD:Facts}
\BIBentryALTinterwordspacing
``{2019 Alzheimer's disease facts and figures},'' \emph{Alzheimer's {\&}
  Dementia}, vol.~15, no.~3, pp. 321--387, 2019. [Online]. Available:
  \url{https://doi.org/10.1016/j.jalz.2019.01.010}
\BIBentrySTDinterwordspacing

\bibitem{AlzheimersAssociation2014}
{Alzheimer's Association}, ``{2014 Alzheimer's disease facts and figures.}''
  \emph{Alzheimer's {\&} dementia : the journal of the Alzheimer's
  Association}, 2014.

\bibitem{NEURO:Gozes17neuroprotectionAD}
I.~Gozes, Ed., \emph{Neuprotection in {Alzheimer}'s disease}.\hskip 1em plus
  0.5em minus 0.4em\relax Amesterdam ; Boston: Elsevier/Academic Press, 2017,
  oCLC: ocn959871992.

\bibitem{NEURO:Kim17neuroprotectionPD}
K.-S. Kim, ``\BIBforeignlanguage{en}{Toward neuroprotective treatments of
  {Parkinson}’s disease},'' \emph{\BIBforeignlanguage{en}{Proceedings of the
  National Academy of Sciences}}, vol. 114, no.~15, pp. 3795--3797, Apr. 2017.

\bibitem{SWMC:Thung16}
K.-H. Thung, E.~Adeli, P.-T. Yap, and D.~Shen, ``Stability-weighted matrix
  completion of incomplete multi-modal data for disease diagnosis,'' in
  \emph{International Conference on Medical Image Computing and
  Computer-Assisted Intervention}, 2016, pp. 88--96.

\bibitem{MedIA2018:Thung2018}
\BIBentryALTinterwordspacing
K.~H. Thung, P.~T. Yap, E.~Adeli, S.~W. Lee, and D.~Shen, ``{Conversion and
  time-to-conversion predictions of mild cognitive impairment using low-rank
  affinity pursuit denoising and matrix completion},'' \emph{Medical Image
  Analysis}, vol.~45, pp. 68--82, 2018. [Online]. Available:
  \url{https://doi.org/10.1016/j.media.2018.01.002}
\BIBentrySTDinterwordspacing

\bibitem{TMI2019:Zhou2019}
T.~Zhou, M.~Liu, K.~H. Thung, and D.~Shen, ``{Latent Representation Learning
  for Alzheimer's Disease Diagnosis With Incomplete Multi-Modality Neuroimaging
  and Genetic Data},'' \emph{IEEE transactions on medical imaging}, vol.~38,
  no.~10, pp. 2411--2422, 2019.

\bibitem{GCN:Bronstein17}
M.~M. Bronstein, J.~Bruna, Y.~Lecun, A.~Szlam, and P.~Vandergheynst,
  ``{Geometric Deep Learning: Going beyond Euclidean data},'' \emph{IEEE Signal
  Processing Magazine}, vol.~34, no.~4, pp. 18--42, 2017.

\bibitem{GCN:Kipf16}
T.~N. Kipf and M.~Welling, ``Semi-supervised classification with graph
  convolutional networks,'' \emph{CoRR}, vol. arXiv:1609.02907, 2016.

\bibitem{GCN:Parisot18MedIA}
S.~Parisot, S.~I. Ktena, E.~Ferrante, M.~Lee, R.~Guerrero, B.~Glocker, and
  D.~Rueckert, ``Disease prediction using graph convolutional networks:
  application to autism spectrum disorder and alzheimer’s disease,''
  \emph{Medical image analysis}, vol.~48, pp. 117--130, 2018.

\bibitem{GCN:Parisot17}
S.~Parisot, S.~I. Ktena, E.~Ferrante, M.~Lee, R.~G. Moreno, B.~Glocker, and
  D.~Rueckert, ``Spectral graph convolutions for population-based disease
  prediction,'' in \emph{International Conference on Medical Image Computing
  and Computer-Assisted Intervention}, 2017, pp. 177--185.

\bibitem{ISBI:Kazi19SelfAtt}
\BIBentryALTinterwordspacing
A.~Kazi, S.~krishna, S.~Shekarforoush, K.~Kortuem, S.~Albarqouni, and N.~Navab,
  ``Self-{Attention} {Equipped} {Graph} {Convolutions} for {Disease}
  {Prediction},'' in \emph{2019 {IEEE} 16th {International} {Symposium} on
  {Biomedical} {Imaging} ({ISBI} 2019)}.\hskip 1em plus 0.5em minus 0.4em\relax
  Venice, Italy: IEEE, Apr. 2019, pp. 1896--1899. [Online]. Available:
  \url{https://ieeexplore.ieee.org/document/8759274/}
\BIBentrySTDinterwordspacing

\bibitem{MICCAI2019:Kazi}
A.~Kazi, S.~Shekarforoush, S.~{Arvind Krishna}, H.~Burwinkel, G.~Vivar,
  B.~Wiestler, K.~Kort{\"{u}}m, S.-A. Ahmadi, S.~Albarqouni, and N.~Navab,
  ``{Graph Convolution Based Attention Model for Personalized Disease
  Prediction},'' in \emph{Medical Image Computing and Computer Assisted
  Intervention -- MICCAI 2019}, D.~Shen, T.~Liu, T.~M. Peters, L.~H. Staib,
  C.~Essert, S.~Zhou, P.-T. Yap, and A.~Khan, Eds.\hskip 1em plus 0.5em minus
  0.4em\relax Cham: Springer International Publishing, 2019, pp. 122--130.

\bibitem{IPMI:Kazi19InceptionGCN}
\BIBentryALTinterwordspacing
A.~Kazi, S.~Shekarforoush, S.~Arvind~Krishna, H.~Burwinkel, G.~Vivar,
  K.~Kortüm, S.-A. Ahmadi, S.~Albarqouni, and N.~Navab,
  ``\BIBforeignlanguage{en}{{InceptionGCN}: {Receptive} {Field} {Aware} {Graph}
  {Convolutional} {Network} for {Disease} {Prediction}},'' in
  \emph{\BIBforeignlanguage{en}{Information {Processing} in {Medical}
  {Imaging}}}, A.~C.~S. Chung, J.~C. Gee, P.~A. Yushkevich, and S.~Bao,
  Eds.\hskip 1em plus 0.5em minus 0.4em\relax Cham: Springer International
  Publishing, 2019, vol. 11492, pp. 73--85. [Online]. Available:
  \url{https://doi.org/10.1007/978-3-030-20351-1_6}
\BIBentrySTDinterwordspacing

\bibitem{Coates2019}
J.~Valenchon and M.~Coates, ``Multiple-graph recurrent graph convolutional
  neural network architectures for predicting disease outcomes,'' in
  \emph{ICASSP 2019-2019 IEEE International Conference on Acoustics, Speech and
  Signal Processing (ICASSP)}.\hskip 1em plus 0.5em minus 0.4em\relax IEEE,
  2019, pp. 3157--3161.

\bibitem{RGCN:Monti17}
F.~Monti, M.~M. Bronstein, and X.~Bresson, ``Geometric matrix completion with
  recurrent multi-graph neural networks,'' \emph{CoRR}, vol. arXiv:1704.06803,
  2017.

\bibitem{Vivar2018}
G.~Vivar, A.~Zwergal, N.~Navab, and S.~A. Ahmadi, ``{Multi-modal disease
  classification in incomplete datasets using geometric matrix completion},''
  in \emph{Lecture Notes in Computer Science (including subseries Lecture Notes
  in Artificial Intelligence and Lecture Notes in Bioinformatics)}, 2018.

\bibitem{atlas_rohlfing2010sri24}
T.~Rohlfing, N.~M. Zahr, E.~V. Sullivan, and A.~Pfefferbaum, ``The sri24
  multichannel atlas of normal adult human brain structure,'' \emph{Human brain
  mapping}, vol.~31, no.~5, pp. 798--819, 2010.

\bibitem{iglesias2011robust}
J.~E. Iglesias, C.-Y. Liu, P.~M. Thompson, and Z.~Tu, ``Robust brain extraction
  across datasets and comparison with publicly available methods,'' \emph{IEEE
  transactions on medical imaging}, vol.~30, no.~9, pp. 1617--1634, 2011.

\bibitem{Preprocessing:baur2018deep}
C.~Baur, B.~Wiestler, S.~Albarqouni, and N.~Navab, ``Deep autoencoding models
  for unsupervised anomaly segmentation in brain mr images,'' in
  \emph{International MICCAI Brainlesion Workshop}.\hskip 1em plus 0.5em minus
  0.4em\relax Springer, 2018, pp. 161--169.

\bibitem{MC:Rao15}
N.~Rao, H.-F. Yu, P.~Ravikumar, and I.~S. Dhillon, ``{Collaborative Filtering
  with Graph Information: Consistency and Scalable Methods},'' \emph{Neural
  Information Processing Systems (NIPS)}, pp. 1--9, 2015.

\bibitem{GMC:Kalofolias14}
V.~Kalofolias, X.~Bresson, M.~Bronstein, and P.~Vandergheynst, ``Matrix
  completion on graphs,'' \emph{arXiv:1408.1717}, 2014.

\bibitem{MC:Goldberg10}
A.~Goldberg, B.~Recht, J.~Xu, R.~Nowak, and X.~Zhu, ``Transduction with matrix
  completion: Three birds with one stone,'' in \emph{Advances in Neural
  Information Processing Systems (NIPS)}, 2010, pp. 757--765.

\bibitem{Vaswani2017}
A.~Vaswani, N.~Shazeer, N.~Parmar, J.~Uszkoreit, L.~Jones, A.~N. Gomez,
  {\L}.~Kaiser, and I.~Polosukhin, ``{Attention is all you need},'' in
  \emph{Advances in Neural Information Processing Systems}, 2017.

\bibitem{GCN:Defferrard16}
M.~Defferrard, X.~Bresson, and P.~Vandergheynst, ``Convolutional neural
  networks on graphs with fast localized spectral filtering,'' in
  \emph{Advances in Neural Information Processing Systems (NIPS)}, 2016, pp.
  3844--3852.

\bibitem{kingma2014adam}
D.~P. Kingma and J.~Ba, ``Adam: A method for stochastic optimization,''
  \emph{arXiv preprint arXiv:1412.6980}, 2014.

\bibitem{paszke2019pytorch}
A.~Paszke, S.~Gross, F.~Massa, A.~Lerer, J.~Bradbury, G.~Chanan, T.~Killeen,
  Z.~Lin, N.~Gimelshein, L.~Antiga \emph{et~al.}, ``Pytorch: An imperative
  style, high-performance deep learning library,'' in \emph{Advances in Neural
  Information Processing Systems}, 2019, pp. 8024--8035.

\bibitem{bergstra2015hyperopt}
J.~Bergstra, B.~Komer, C.~Eliasmith, D.~Yamins, and D.~D. Cox, ``Hyperopt: a
  python library for model selection and hyperparameter optimization,''
  \emph{Computational Science \& Discovery}, vol.~8, no.~1, p. 014008, 2015.

\bibitem{MODEL:breiman2001randomforest}
L.~Breiman, ``Random forests,'' \emph{Machine learning}, vol.~45, no.~1, pp.
  5--32, 2001.

\bibitem{troyanskaya2001KNN_imputation}
O.~Troyanskaya, M.~Cantor, G.~Sherlock, P.~Brown, T.~Hastie, R.~Tibshirani,
  D.~Botstein, and R.~B. Altman, ``Missing value estimation methods for dna
  microarrays,'' \emph{Bioinformatics}, vol.~17, no.~6, pp. 520--525, 2001.

\bibitem{buuren2010mice}
S.~v. Buuren and K.~Groothuis-Oudshoorn, ``mice: Multivariate imputation by
  chained equations in r,'' \emph{Journal of statistical software}, pp. 1--68,
  2010.

\bibitem{pedregosa2011scikit}
F.~Pedregosa, G.~Varoquaux, A.~Gramfort, V.~Michel, B.~Thirion, O.~Grisel,
  M.~Blondel, P.~Prettenhofer, R.~Weiss, V.~Dubourg \emph{et~al.},
  ``Scikit-learn: Machine learning in python,'' \emph{Journal of machine
  learning research}, vol.~12, no. Oct, pp. 2825--2830, 2011.

\bibitem{komer2014hyperopt-sklearn}
B.~Komer, J.~Bergstra, and C.~Eliasmith, ``Hyperopt-sklearn: automatic
  hyperparameter configuration for scikit-learn,'' in \emph{ICML workshop on
  AutoML}, vol.~9.\hskip 1em plus 0.5em minus 0.4em\relax Citeseer, 2014.

\bibitem{NeuroImage2013:RandomForest_Gray}
K.~R. Gray, P.~Aljabar, R.~A. Heckemann, A.~Hammers, D.~Rueckert, A.~D.~N.
  Initiative \emph{et~al.}, ``Random forest-based similarity measures for
  multi-modal classification of alzheimer's disease,'' \emph{NeuroImage},
  vol.~65, pp. 167--175, 2013.

\bibitem{langa2014MCIADbaseline}
K.~M. Langa and D.~A. Levine, ``The {Diagnosis} and {Management} of {Mild}
  {Cognitive} {Impairment}: {A} {Clinical} {Review},'' \emph{JAMA}, vol. 312,
  no.~23, p. 2551, 2014.

\bibitem{hamilton2017graphsage}
W.~Hamilton, Z.~Ying, and J.~Leskovec, ``Inductive representation learning on
  large graphs,'' in \emph{Advances in neural information processing systems},
  2017, pp. 1024--1034.

\bibitem{velickovic2018gat}
\BIBentryALTinterwordspacing
P.~Veli{\v{c}}kovi{\'{c}}, G.~Cucurull, A.~Casanova, A.~Romero, P.~Li{\`{o}},
  and Y.~Bengio, ``{Graph Attention Networks},'' \emph{International Conference
  on Learning Representations}, 2018. [Online]. Available:
  \url{https://openreview.net/forum?id=rJXMpikCZ}
\BIBentrySTDinterwordspacing

\bibitem{wang2016tsvm_based_for_missing_data}
G.~Wang, Z.~Deng, and K.-S. Choi, ``Tackling missing data in community health
  studies using additive ls-svm classifier,'' \emph{IEEE journal of biomedical
  and health informatics}, vol.~22, no.~2, pp. 579--587, 2016.

\bibitem{venugopalan2019clustering_based_for_missing_data}
J.~Venugopalan, N.~Chanani, K.~Maher, and M.~D. Wang, ``Novel data imputation
  for multiple types of missing data in intensive care units,'' \emph{IEEE
  journal of biomedical and health informatics}, vol.~23, no.~3, pp.
  1243--1250, 2019.

\end{thebibliography}

\end{document}